%% file: main.tex
\documentclass[letterpaper, 10 pt, conference]{ieeeconf}
\IEEEoverridecommandlockouts 
\usepackage[letterpaper, left=54pt, right=54pt, bottom=54pt, top=54pt]{geometry}

\usepackage{graphicx}
\usepackage{amsmath} 
\usepackage{amssymb}  
\usepackage{mathtools}
\usepackage{bbm}
\usepackage{dsfont}
\usepackage{multirow}
\usepackage[dvipsnames]{xcolor}
\usepackage{subfig}
\usepackage[noadjust]{cite}
\usepackage{enumerate}
\usepackage{booktabs}
\usepackage{hyperref}

\usepackage{algpseudocode}
\usepackage{algorithm}
\usepackage{caption}
\usepackage[free-standing-units=true]{siunitx}
\usepackage{cleveref}

\title{\LARGE \bf
Neural Informed RRT*: Learning-based Path Planning with Point Cloud State Representations under Admissible Ellipsoidal Constraints}

\author{Zhe Huang, Hongyu Chen, John Pohovey, and Katherine Driggs-Campbell
\thanks{Z. Huang, H. Chen, J. Pohovey, and K. Driggs-Campbell are with the Department of  Electrical and Computer Engineering at the University of Illinois at Urbana-Champaign. emails: \{zheh4, hongyuc5, jpohov2, krdc\}@illinois.edu}%
\thanks{
This work was supported by the National Science Foundation under Grant No. 2143435.
}
}

\begin{document}

\maketitle
\thispagestyle{empty}
\pagestyle{empty}

\input{00_Abstract}
\input{01_Introduction}
\input{02_Related_Work}
\input{03_Method}
\input{04_Experiments}
\input{05_Conclusions}

\bibliographystyle{IEEEtran}
\bibliography{bib}

\end{document}

%% file: 00_Abstract.tex
 \begin{abstract}
    Sampling-based planning algorithms like Rapidly-exploring Random Tree (RRT) are versatile in solving path planning problems. RRT* offers asymptotic optimality but requires growing the tree uniformly over the free space, which leaves room for efficiency improvement. To accelerate convergence, rule-based informed approaches sample states in an admissible ellipsoidal subset of the space determined by the current path cost. Learning-based alternatives model the topology of the free space and infer the states close to the optimal path to guide planning. We propose Neural Informed RRT* to combine the strengths from both sides. We define point cloud representations of free states. We perform \textit{Neural Focus}, which constrains the point cloud within the admissible ellipsoidal subset from Informed RRT*, and feeds into PointNet++ for refined guidance state inference. In addition, we introduce \textit{Neural Connect} to build connectivity of the guidance state set and further boost performance in challenging planning problems. Our method surpasses previous works in path planning benchmarks while preserving probabilistic completeness and asymptotic optimality. We deploy our method on a mobile robot and demonstrate real world navigation around static obstacles and dynamic humans. Code is available at \url{https://github.com/tedhuang96/nirrt_star}.
\end{abstract}

%% file: 01_Introduction.tex
\section{Introduction}
Path planning is the task of finding a path for a robot to traverse from a start to a goal safely and efficiently~\cite{lavalle2006planning, zhang2018path, patle2019review}. An effective path planning algorithm should be (1) complete and optimal: a solution is guaranteed to be found if one exists, and the optimal solution is guaranteed to be achieved with sufficient run time; (2) efficient in optimal convergence: the solution should be quickly improved towards near optimal; and (3) versatile and scalable: the implementation should be modified with minimal effort to generalize across different problems, environments, and robots.
\begin{figure}[t]
    \centering\includegraphics[width=0.9\linewidth]{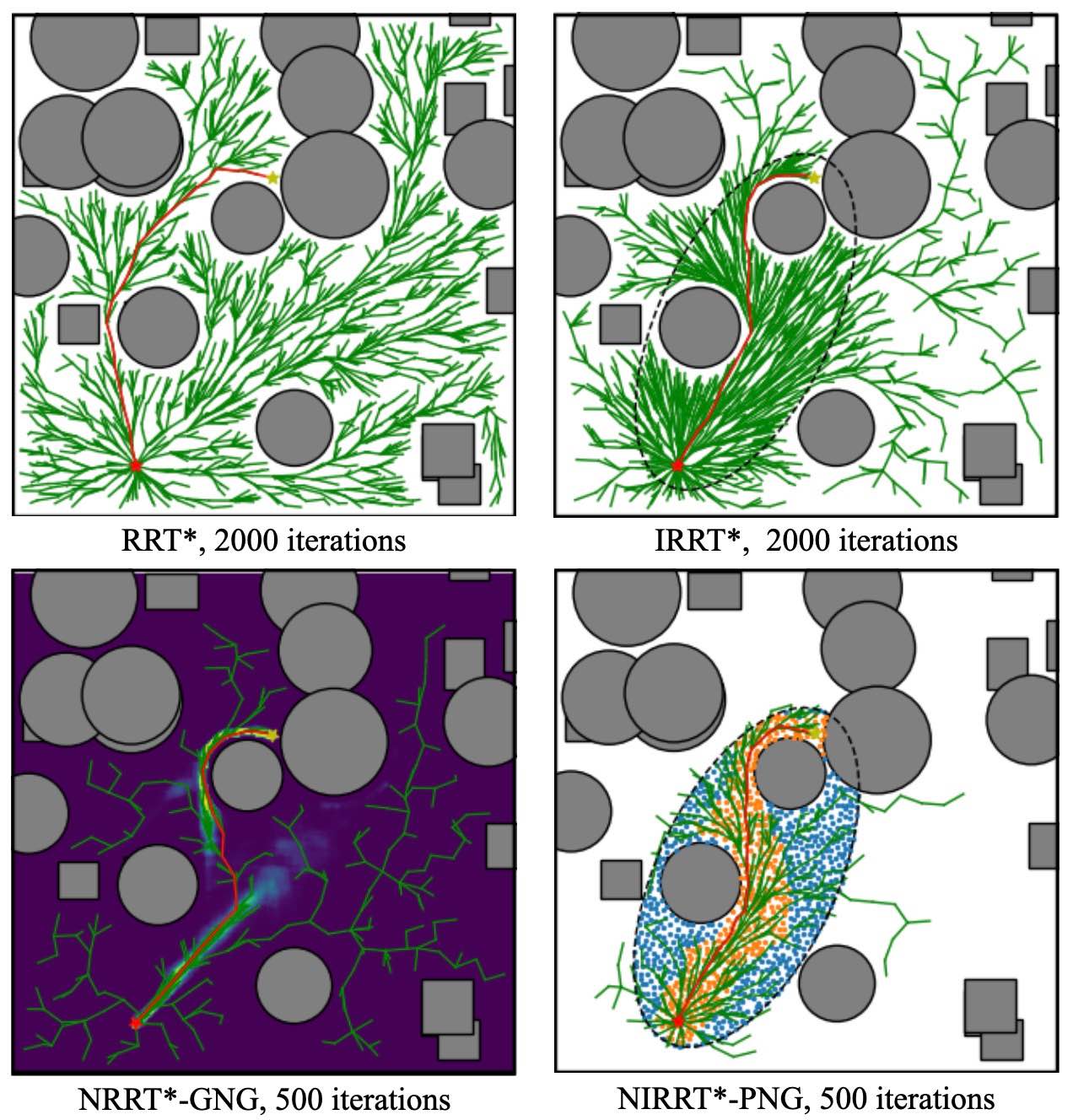}
    \caption{Solutions of a 2D random world found by RRT*~\cite{karaman2011sampling}, Informed RRT* (IRRT*)~\cite{gammell2014informed}, Neural RRT* with Grid-based Network Guidance (NRRT*-GNG)~\cite{wang2020neural}, and Neural Informed RRT* with Point-based Network Guidance (NIRRT*-PNG). NIRRT* effectively integrates IRRT* and point-based network, so IRRT* helps point-based network focus guidance state inference on the important region for solution improvement, and point-based network helps IRRT* sample critical states in the admissible ellipsoidal subset for convergence acceleration.}
    \vspace{-20pt}
    \label{fig:intro}
\end{figure}

Multiple branches of planning algorithms have been developed to meet these requirements, including grid-based search, artificial potential field, and sampling-based algorithms~\cite{hart1968formal, stentz1994optimal, khatib1986real, kavraki1996probabilistic, lavalle2001randomized}. Sampling-based algorithms are popular due to their versatility, scalability, and formal properties of probabilistic completeness and asymptotic optimality~\cite{karaman2011sampling}. To accelerate convergence to the optimal path, various sampling strategies are introduced to replace the default uniform sampling~\cite{karaman2011sampling}. The rule-based informed strategy enforces sampling in an admissible ellipsoidal subset of states which are more promising to improve the current path solution~\cite{gammell2014informed, gammell2015batch}. 
The learning-based methods harness grid-based neural networks to make inference of states close to the optimal path, and bias sampling towards these states, which we define as guidance states~\cite{perez2018learning, wang2020neural, ichter2020learned, qureshi2020motion, bency2019neural, kumar2019lego, ichter2018learning, ma2021conditional}.

While these works improve performance, we observe three limitations. First, learning-based methods encode whole state space to generate guidance states without iterative improvement, where inference speed and accuracy are affected by modeling features of irrelevant region or obstacles. Second, rule-based informed sampling does not favor topologically critical states in the ellipsoidal subset (e.g., narrow corridors). Finally, learning-based methods do not consider connectivity of the guidance state set, which severely affects the convergence rate in complex planning problems.

We introduce Neural Informed RRT* (NIRRT*) to address these limitations (Figure~\ref{fig:intro}). We represent free states with a point cloud, and apply PointNet++~\cite{qi2017pointnet++} to classify guidance states. Sampling from the guidance states is mixed with the random sampling step of Informed RRT* (IRRT*)~\cite{gammell2014informed}. Using a point cloud instead of an occupancy grid allows us to perform \textit{Neural Focus}: constraining the point clouds by the admissible ellipsoidal subset of the free space, from which the critical states are inferred by PointNet++. The quality of the guidance states is continually improved during iteration, because the inference is always made on the informed subset created by an improved path cost. In addition, we build connectivity of the guidance state set by following a \textit{Neural Connect} scheme similar to RRT-Connect~\cite{kuffner2000rrt}, where the point-based network is called to solve a subproblem with a closer pair of start and goal states.

In short, our contributions are threefold: (1) we use a Point-based Network (PointNet++) to directly take free states as point cloud input to generate multiple guidance states in one run; (2) we present Neural Informed RRT*, by introducing Neural Focus to integrate Point-based Network and Informed RRT*; and (3) we propose Neural Connect to address the connectivity issue of inferred guidance state set.

%% file: 02_Related_Work.tex
\section{Related Work}\label{sec:related-work}
Grid-based search methods like A*~\cite{hart1968formal} and D*~\cite{stentz1994optimal} guarantee to find the optimal path in a discretized state space if a solution exists, at the cost of poor scaling with the problem complexity. Sampling-based algorithms like probabilistic roadmap (PRM)~\cite{kavraki1996probabilistic} and RRT~\cite{lavalle2001randomized} guarantee to find a feasible path solution if one exists as the number of iterations approaches infinity. PRM* and RRT*~\cite{karaman2011sampling} provide asymptotic optimality, which requires exploring the planning domain globally. Informed RRT* and Batch Informed Trees improve the convergence rate by constraining the sampling space to a ellipsoidal subset based on start state, goal state, and current best path cost~\cite{gammell2014informed, gammell2015batch}.

Another line of works accelerates path planning by investigating the search space of the problem, such as Vonoroi bias~\cite{yershova2005dynamic, wang2020optimal}, evolutionary algorithms~\cite{martin2007offline}, and A* initialization~\cite{brunner2013hierarchical}. Neural RRT* represents the square-shaped search space of 2D planning problems by images and uses U-Net~\cite{ronneberger2015u} to predict a probabilistic heatmap of states used for guiding RRT*~\cite{wang2020neural}. MPNet voxelizes environment point cloud and feed into 3D Convolution Neural Networks to make recursive inference for path generation~\cite{qureshi2020motion}. Grid-based neural networks are prevalent in previous works to encode the search space~\cite{perez2018learning,wang2020neural,qureshi2020motion,ichter2020learned, ma2021conditional}, which requires discretization operations and the results are dependent on resolution. Previous works use PointNet to encode the point cloud of obstacle states~\cite{strudel2021learning, sugiura2022p3net}, but modeling the obstacle interior is inefficient for finding a path in the free space. Recent works apply graph neural networks to a sampled random geometric graph in configuration space, and select edges from the graph to build a near-optimal path~\cite{yu2021reducing, zhang2022learning}. However, the path feasibility and path quality are highly dependent on the sampled graph, while continuous improvement is not discussed in these works.

%% file: 03_Method.tex
\section{Method}\label{sec:method}
\subsection{Problem Definition}
We define the optimal path planning problem similar to related works~\cite{karaman2011sampling, gammell2014informed, wang2020neural}. The state space is denoted as $X \subseteq \mathbb{R}^{d}$. The obstacle space and the free space are denoted as $X_{\textrm{obs}}$ and $X_{\textrm{free}}$. A path $\sigma: [0, 1] \rightarrow X_{\textrm{free}}$ is a sequence of states. The set of paths is denoted as $\Sigma$. The optimal path planning problem is to find a path $\sigma^{*}$ which minimizes a given cost function $c: \Sigma \rightarrow \mathbb{R}_{\geq 0}$, connects a given start state $x_{\textrm{start}} \in X_{\textrm{free}}$ and a given goal state $x_{goal} \in X_{\textrm{free}}$, and has all states on the path in free space. Formally:
\begin{equation}
\begin{aligned}
    \sigma^{*} = & \arg \min_{\sigma \in \Sigma} c\left(\sigma\right) \\ \textrm{s.t. }  \sigma(0)=x_{\textrm{start}}, \sigma(1)= & x_{\textrm{goal}}, \forall s \in [0, 1], \sigma(s) \in X_{\textrm{free}} \\
\end{aligned}
\end{equation}

\subsection{Neural Informed RRT*}
We present NIRRT* in Algorithm~\ref{algo:NIRRT*-PNG}, where the unhighlighted part is from RRT*, the blue part is from IRRT*, and the red part is our contribution. We track the best path solution cost $c_{\textrm{best}}^i$ through each iteration, which is initialized as infinity (line~\ref{algo:NIRRT*-PNG-l3}). We initialize update cost $c_{\textrm{update}}$ with the value of $c_{\textrm{best}}^0$ (line~\ref{algo:NIRRT*-PNG-l4}). We call the neural network to infer an initial guidance state set $X_{\textrm{guide}}$ based on the complete free state space (line~\ref{algo:NIRRT*-PNG-l5}). As better solutions are found, the guidance state set $X_{\textrm{guide}}$ may be updated by the neural network calls depending on how much the path cost has been improved, and random samples $x_{\textrm{rand}}$ are sampled using both $X_{\textrm{guide}}$ and informed sampling (line~\ref{algo:NIRRT*-PNG-l8}).

\input{NeuralInformedRRT}

\input{PointNetGuidedSampling}

$\texttt{PointNetGuidedSampling}$: When the current best path cost $c_{\textrm{curr}}$ is less than the path cost improvement ratio $\alpha \leq 1$ of $c_{\textrm{update}}$, the neural network is called to update $X_\textrm{guide}$, and $c_{\textrm{update}}$ is updated by $c_{\textrm{curr}}$. The random sample $x_{\textrm{rand}}$ is sampled with a mixed strategy: if a random number $\textrm{Rand()} \in (0, 1)$ is smaller than 0.5, we use the sampling strategy of IRRT* to sample $x_{\textrm{rand}}$; otherwise, we sample $x_{\textrm{rand}}$ uniformly from $X_{\textrm{guide}}$. Similar to~\cite{wang2020neural,qureshi2020motion,ichter2018learning}, our mixed sampling strategy guarantees probabilistic completeness and asymptotic optimality by implementing the sampling procedure of IRRT* with a non-zero probability.

Note the frequency of calling neural networks for guidance state inference is controlled by the path cost improvement ratio $\alpha$. If we do not update $X_{\textrm{guide}}$ after initial inference, and remove IRRT* components, NIRRT* is reduced to NRRT*. While NIRRT* is generic in that any neural network that infers guidance states can fit into the framework, we emphasize the use of a point-based network. In the next subsection, we discuss the details of Point-based Network Guidance (PNG), and explain the preference of point representations over grid representations.

\subsection{Point-based Network Guidance}
\label{sec:method-point-based-network-guidance} 
\textbf{Point-based Network.} We represent the state space by a point cloud $X_{\textrm{input}} = \{x_{1}, x_{2}, \ldots, x_{N}\} \subset X_{\textrm{free}}$. The density of point cloud should allow a reasonable amount of neighbors around each point in radius of step size $\eta$. We oversample points uniformly from $X_{\textrm{free}}$, and perform minimum distance downsampling to obtain the point cloud with even distribution. We create a one-hot vector for each point, indicating whether the point is within radius $\eta$ of $x_{\textrm{start}}$ or $x_{\textrm{goal}}$. We concatenate the one-hot vectors with normalized point coordinates to generate point cloud representations of the free states. The processed point cloud $\Bar{X}_{\textrm{input}}$ is fed into a point-based network~$f$. The network~$f$ maps each point to a probability $p_{i} \in [0, 1]$, where the points with probability greater than 0.5 form the set of guidance states $X_{\textrm{guide}}$. Formally,
\begin{equation}
    \{p_{1}, p_{2}, \ldots, p_{N}\} = f(\Bar{X}_{\textrm{input}}), \,\, X_{\textrm{guide}} = \{x_{i} | p_{i} > 0.5\}.
\end{equation}

We implement PointNet++~\cite{qi2017pointnet++} as the model architecture of the point-based network. Since PointNet++ is originally designed for 3D point cloud, we set $z$ coordinates as zero for 2D problems. We collect 4,000 2D random worlds as the training dataset. For each random world, we run A* in pixel space with step size of unit pixel and clearance of 3 pixels to generate the pixel-wise optimal path. We generate a point cloud of number $N=2048$, and generate guidance state labels by checking whether each point is around any point of the pixel-wise optimal path in radius of $\eta$, which is set as 10 pixels. We train PointNet++ by Adam optimizer~\cite{kingma2015adam} with an initial learning rate of 0.001 and batch size of 16 for 100 epochs. We use the trained model across all types of 2D planning problems. For 3D random world problems, we follow a similar scheme, but the clearance is set as 2 voxels.

\begin{figure*}[hbt!]
    \centering
    \includegraphics[width=0.88\linewidth]{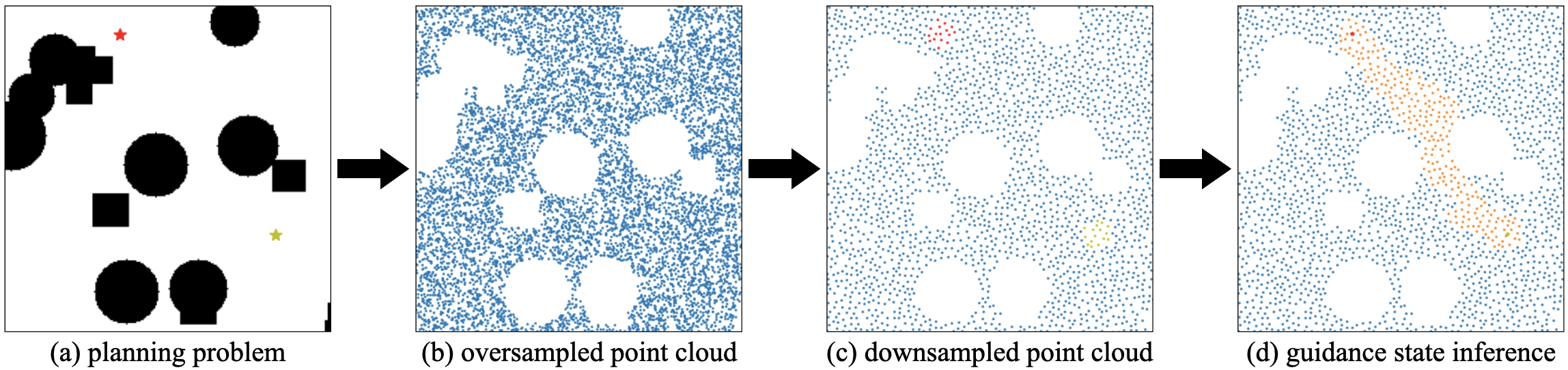}
    \caption{Guidance state inference by point-based network. Red is start, yellow is goal, blue is free states, and orange is guidance states.}
    \label{fig:point-based-network}
\end{figure*}
\begin{figure*}[hbt!]
    \centering
    \includegraphics[width=0.92\linewidth]{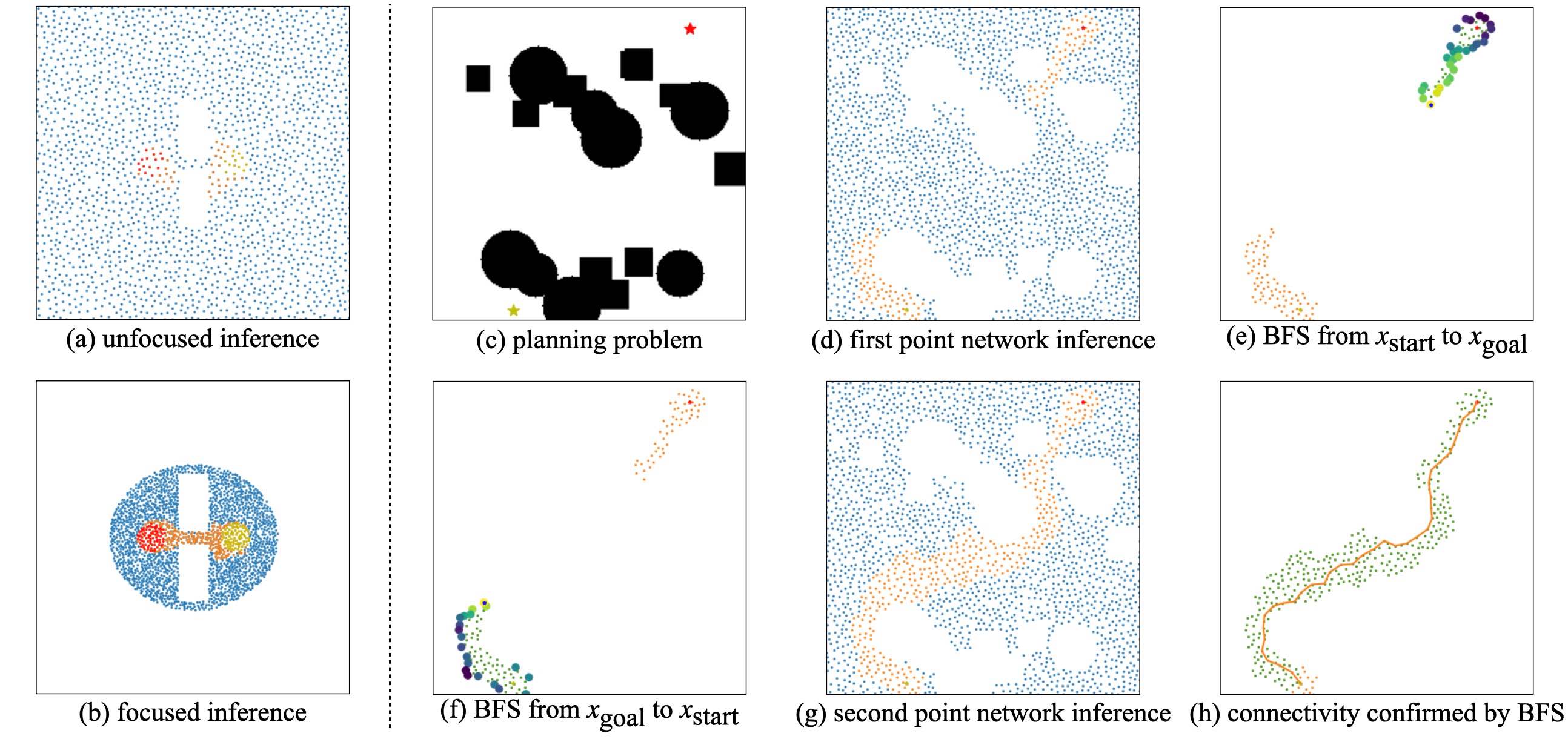}
    \caption{(a-b) Neural Focus in a 2D narrow passage and (c-h) Neural Connect in a 2D random world. Green dots denote states visited by Breadth First Search (BFS). Circles of larger size around green dots denote boundary points. The colors of circles denote the heuristic scores, where brighter colors represent higher scores. The boundary point which is $x_{\textrm{start}}^{i+1}$ or $x_{\textrm{goal}}^{i+1}$ has a blue marker on the circle. The orange line denotes the path found by BFS which represents the connectivity of $X_{\textrm{guide}}$.}
    \label{fig:neural-connect-focus}
    \vspace{-20pt}
\end{figure*}

\textbf{Neural Focus.} Informed RRT* outperforms RRT* by proposing a heuristic ellipsoidal subset of the planning domain $X_{\textrm{focus}}$ in terms of the current best solution cost $c_{\textrm{curr}}$, in order to sample $x_{\textrm{rand}}$ which is more likely to improve the current solution. The reasoning behind this sampling strategy is that for any state $x_{\textrm{rejected}}$ from $X \backslash X_{\textrm{focus}}$, the minimum cost of a feasible path from $x_{\textrm{start}}$ to $x_{\textrm{goal}}$ through $x_{\textrm{rejected}}$ is greater than $c_{\textrm{curr}}$:
\begin{equation}
    X_{\textrm{focus}} = \{x \in X \big|\, ||x-x_{\textrm{start}}||_2+||x-x_{\textrm{goal}}||_2 \leq c_{\textrm{curr}}\}
\end{equation}
 
Neural Focus is to constrain the point cloud input to the point-based network inside the $X_{\textrm{focus}}$, which is equivalent as changing the domain of oversampling from $X_{\textrm{free}}$ to $X_{\textrm{focus}}\cap X_{\textrm{free}}$. Since we normalize point coordinates when processing point cloud inputs, the trained point-based network can handle point clouds sampled from domains at different scales. With the same number of points $N$, a smaller volume of $X_{\textrm{focus}}$ leads to a denser point cloud, which describes important regions with finer details. For example, Figure~\ref{fig:neural-connect-focus}(b) shows that Neural Focus fills the narrow passage with a large number of points, which is captured by the point-based network to produce more effective inference on guidance states compared to Figure~\ref{fig:neural-connect-focus}(a). 

\textbf{Neural Connect.} The points close to $x_{\textrm{start}}$ or $x_{\textrm{goal}}$ are usually classified as guidance states with greater probabilities than the points around midway of the path (e.g., Figure~\ref{fig:neural-connect-focus}(d)). When the distance between $x_{\textrm{start}}$ and $x_{\textrm{goal}}$ gets longer, the guidance state set $X_{\textrm{guide}}$ is more likely to be separated into disconnected ``blobs''. This phenomenon of probability polarization is reported in NRRT* work~\cite{wang2020neural}. Our experiments show lack of connectivity limits the performance in large and complex planning problems. 

We address this issue by introducing Neural Connect, which is inspired by RRT-Connect~\cite{kuffner2000rrt}. We initialize $X_{\textrm{guide}}$ as an empty set, $x_{\textrm{start}}^1$ as $x_{\textrm{start}}$, and $x_{\textrm{goal}}^1$ as $x_{\textrm{goal}}$. During iteration, we first call the point-based network with $x_{\textrm{start}}^i$ and $x_{\textrm{goal}}^i$ as start and goal, and add inferred guidance states to $X_{\textrm{guide}}$. Second, We run Breadth First Search (BFS) from $x_{\textrm{start}}$ to $x_{\textrm{goal}}$ through the guidance states in $X_{\textrm{guide}}$. The neighbor radius of BFS is set as $\eta$, and no collision check is performed. After BFS is finished, connectivity of $X_{\textrm{guide}}$ is confirmed if $x_{\textrm{goal}}$ is reached. Otherwise, we find the boundary points $X_{\textrm{bound}}$ of the states visited by BFS by checking whether any points in $X_{\textrm{input}} \backslash X_{\textrm{guide}}$ are around the visited state of radius $\eta/2$. We select $x_{\textrm{start}}^{i+1}$ from $X_{\textrm{bound}}$ which is one of the states heuristically the furthest from $x_{\textrm{start}}$ and one of the states to reach $x_{\textrm{goal}}$ with minimum total heuristic cost. Third, we perform the same operation as the second step, with the start of BFS as $x_{\textrm{goal}}$, and the goal of BFS as $x_{\textrm{start}}$. We obtain $x_{\textrm{goal}}^{i+1}$ if connectivity is negative. We perform the iteration until connectivity is built or the limit of iteration $n_{\textrm{guide}}$ is reached, which we set as 5 in practice. We illustrate Neural Connect in Figure~\ref{fig:neural-connect-focus}(c-h). Note the orange path found by BFS in Figure~\ref{fig:neural-connect-focus}(h) does not go through collision check, so the path is not a feasible solution but a visual demonstration on the connectivity of $X_{\textrm{guide}}$. 

\texttt{PointNetGuide}: We apply both Neural Focus and Neural Connect to the point-based network, and obtain the complete module of Point-based Network Guidance, which is presented in Algorithm~\ref{algo:PointNetGuide}.

\input{PointNetGuide}

\textbf{Point versus Grid.} We prefer using points over grids to represent state space due to compatibility with geometric constraints and convenience of extension to different problems. To apply Neural Focus to a CNN, grid representations require masking of the complement set of the ellipsoidal subset, where the mask quality depends on grid resolution. CNN also has to process the irrelevant masked region within the rectangular/box grid input. In contrast, point representations naturally confine states within arbitrary geometry by modifying the sampling domain, and the point-based network only needs to model free states. Moreover, while the point-based network just needs adjustment of the input format to extend to different dimensions, changing input dimensions usually requires redesign of CNN architecture.

%% file: NeuralInformedRRT.tex
\begin{algorithm}[t!]
    \caption{Neural Informed RRT*}\label{algo:NIRRT*-PNG}
    \begin{algorithmic}[1]
    \State $V \gets \{x_{\textrm{start}}\}; E \gets \varnothing$;
    \color{blue}
    \State $X_{\textrm{soln}} \gets \varnothing$;
    \color{red}
    \State $c_{\textrm{best}}^0 \gets \infty$;\label{algo:NIRRT*-PNG-l3}
    \State $c_{\textrm{update}} \gets c_{\textrm{best}}^0$;\label{algo:NIRRT*-PNG-l4}
    \State $X_{\textrm{guide}} \gets \texttt{PointNetGuide}(x_{\textrm{start}}, x_{\textrm{goal}}, c_{\textrm{best}}^0, X_{\textrm{free}})$;\label{algo:NIRRT*-PNG-l5}
    \color{black}
    \For {$i = 1$ to $n$}
        \color{blue}
        \State $c_{\textrm{best}}^i \gets \min_{x_{\textrm{soln}} \in X_{\textrm{soln}}} \{\texttt{Cost}(x_{\textrm{soln}})\}$;       
        \color{red}
        \State $x_{\textrm{rand}},\!X_{\textrm{guide}},\!c_{\textrm{update}}\!\!\!\gets\!\!\!\texttt{PointNetGuidedSampling}$ $(X_{\textrm{guide}}, x_{\textrm{start}}, x_{\textrm{goal}}, c_{\textrm{update}}, c_{\textrm{best}}^{i}, X_{\textrm{free}})$;\label{algo:NIRRT*-PNG-l8}
        \color{black}
        \State $x_{\textrm{nearest}} \gets \texttt{Nearest}(G=(V,E), x_{\textrm{rand}})$;
        \State $x_{\textrm{new}} \gets \texttt{Steer}(x_{\textrm{nearest}}, x_{\textrm{rand}})$;
        \If{$\texttt{CollisionFree}(x_{\textrm{nearest}}, x_{\textrm{new}})$}
            \State $X_{\textrm{near}} \gets \texttt{Near}(G=(V,E), x_{\textrm{new}}, r_{\textrm{RRT}^*})$;
            \State $V \gets V \cup \{x_{\textrm{new}}\}$;
            \State $x_{\textrm{min}} \gets x_{\textrm{nearest}}$; \State $c_{\textrm{min}} \gets \texttt{Cost}(x_{\textrm{nearest}})+c(\texttt{Line}(x_{\textrm{nearest}}, x_{\textrm{new}}))$;
            \ForAll {$x_{\textrm{near}} \in X_{\textrm{near}}$}
                \If {\!$\texttt{CollisionFree}(\!x_{\textrm{near}},\!x_{\textrm{new}}\!)\!\land\!\texttt{Cost}(\!x_{\textrm{near}}\!)$ $+c(\texttt{Line}(x_{\textrm{near}}, x_{\textrm{new}}))\!<\!c_{\textrm{min}}$}
                    \State $x_{\textrm{min}} \gets x_{\textrm{near}}$;
                    \State $c_{\textrm{min}}\!\gets\!\texttt{Cost}(x_{\textrm{near}})\!+\!c(\texttt{Line}(x_{\textrm{near}},\!x_{\textrm{new}}))$;
                \EndIf
            \EndFor
            \State $E \gets E \cup \{(x_{\textrm{min}}, x_{\textrm{new}})\}$;
            \ForAll {$x_{\textrm{near}} \in X_{\textrm{near}}$}
                \If {$\texttt{CollisionFree}(x_{\textrm{new}},\!x_{\textrm{near}})\!\land\!\texttt{Cost}(\!x_{\textrm{new}}\!)$ $+c(\texttt{Line}(x_{\textrm{new}}, x_{\textrm{near}}))\!<\! \texttt{Cost}(x_{\textrm{near}})$}
                    \State $x_{\textrm{parent}} \gets \texttt{Parent}(x_{\textrm{near}})$;
                    \State $E\!\gets\!(\!E \backslash \{(x_{\textrm{parent}},\!x_{\textrm{near}})\}\!)\cup\{(x_{\textrm{new}},\!x_{\textrm{near}})\}$;
                \EndIf
            \EndFor
            \color{blue}
            \If {$\texttt{InGoalRegion}(x_{\textrm{new}})$}
                \State $X_{\textrm{soln}} \gets X_{\textrm{soln}} \cup \{x_{\textrm{new}}\};$
            \EndIf
            \color{black}
        \EndIf
    \EndFor
    \State \Return G=(V, E);
\end{algorithmic} 
\end{algorithm}

%% file: PointNetGuidedSampling.tex
\begin{algorithm}[t!]
\caption{$\texttt{PointNetGuidedSampling}(X_{\textrm{guide}}, x_{\textrm{start}},$ $ x_{\textrm{goal}}, c_{\textrm{update}}, c_{\textrm{curr}}, X_{\textrm{free}})$}\label{algo:PNGS}
    \begin{algorithmic}[1]
    \If {$c_{\textrm{curr}} < \alpha\, c_{\textrm{update}}$}\label{algo:PNGS:l1}
        \State $X_{\textrm{guide}} \gets \texttt{PointNetGuide}(x_{\textrm{start}}, x_{\textrm{goal}}, c_{\textrm{curr}}, X_{\textrm{free}})$;\label{algo:PNGS:l2}
        \State $c_{\textrm{update}} \gets c_{\textrm{curr}}$;\label{algo:PNGS:l3}
    \EndIf
    \If {$\textrm{Rand()} < 0.5$}\label{algo:PNGS:l5}
        \If {$c_{\textrm{curr}} < \infty$}
            \State $x_{\textrm{rand}} \!\gets\!\texttt{InformedSampling}(x_{\textrm{start}}, x_{\textrm{goal}}, c_{\textrm{curr}})$;
        \Else
            \State $x_{\textrm{rand}}\!\gets\!\texttt{UniformSampling}(X_{\textrm{free}})$;
        \EndIf
    \Else
        \State $x_{\textrm{rand}} \gets \texttt{UniformSampling}(X_{\textrm{guide}})$;
    \EndIf\label{algo:PNGS:l13}
    \State \Return $x_{\textrm{rand}}, X_{\textrm{guide}}, c_{\textrm{update}}$;
\end{algorithmic}
\end{algorithm}

%% file: PointNetGuide.tex
\begin{algorithm}[t!]
    \caption{$\texttt{PointNetGuide}(x_{\textrm{start}}, x_{\textrm{goal}}, c_{\textrm{curr}}, X_{\textrm{free}})$}\label{algo:PointNetGuide}
    \begin{algorithmic}[1]
    \State $x_{\textrm{start}}^1\gets x_{\textrm{start}}$;
    \State $x_{\textrm{goal}}^1 \gets x_{\textrm{goal}}$;
    \State $X_{\textrm{guide}} \gets \varnothing$;
    \If {$c_{\textrm{curr}} < \infty$}
        \State $X_{\textrm{focus}}\!\gets\!\texttt{InformedSubset}(x_{\textrm{start}}, x_{\textrm{goal}}, c_{\textrm{curr}});$
        \State $X_{\textrm{input}}\!\gets\!\texttt{PointCloudSampling}(X_{\textrm{focus}} \cap X_{\textrm{free}})$;
    \Else
        \State $X_{\textrm{input}}\!\gets\!\texttt{PointCloudSampling}(X_{\textrm{free}})$;
    \EndIf
    \For{$j=1$ to $n_{\textrm{guide}}$}
        \State $\Bar{X}_{\textrm{input}}\!\gets\!\texttt{AddOneHotFeatures}(X_{\textrm{input}},\!x_{\textrm{start}}^j,\!x_{\textrm{goal}}^j)$;
        \State $\Bar{X}_{\textrm{input}}\!\gets\!\texttt{NormalizeCoordinates}(\Bar{X}_{\textrm{input}})$;
        \State $X_{\textrm{guide}}\!\gets\!X_{\textrm{guide}} \cup \textrm{PointBasedNetwork}(\Bar{X}_{\textrm{input}})$;
        \State connectivity, $x_{\textrm{start}}^{j+1} \gets \texttt{BFS}(X_{\textrm{guide}}, x_{\textrm{start}}, x_{\textrm{goal}}, \eta)$;
        \If{connectivity}
            \State \Return $X_{\textrm{guide}}$;
        \EndIf
        \State connectivity, $x_{\textrm{goal}}^{j+1}\gets \texttt{BFS}(X_{\textrm{guide}}, x_{\textrm{goal}}, x_{\textrm{start}}, \eta)$;
        \If{connectivity}
            \State \Return $X_{\textrm{guide}}$;
        \EndIf
    \EndFor
    \State \Return $X_{\textrm{guide}}$;
\end{algorithmic} 
\end{algorithm}

%% file: 04_Experiments.tex
\section{Experiments}\label{sec:experiments}
\subsection{Simulation Experiments}
\begin{figure}[hbt!]
    \centering
    \includegraphics[width=0.8\linewidth]{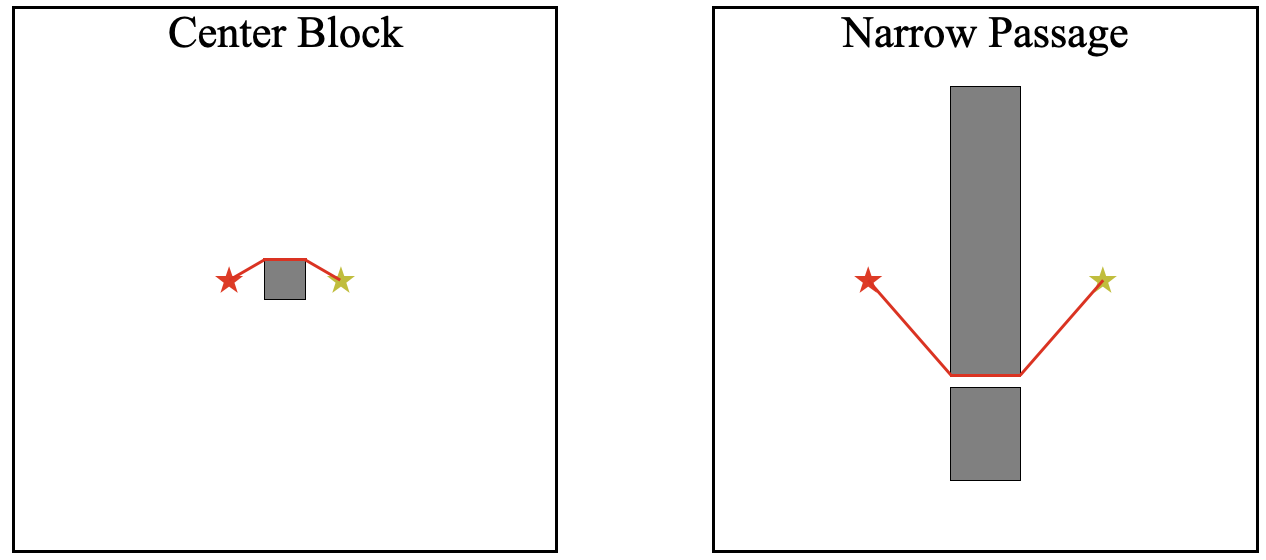}
    \caption{Center block and narrow passage experiments.}
    \label{fig:block_gap_explain}
    \vspace{-20pt}
\end{figure}

\textbf{Planning Problems.} We conduct simulation experiments on 2D center block, 2D narrow passage, 2D random world, and 3D random world problems. The center block and the narrow passage problems are defined similar to IRRT* work~\cite{gammell2014informed} (Figure~\ref{fig:block_gap_explain}). The center block problem examines the efficiency of planners to sample states relevant to the problem in a wide free space. The narrow passage problem studies the capability of planners to focus sampling in topologically critical area. The random world problems evaluate versatility and scalability of planners.

In the center block problems, we specify 5 different map sizes with respect to a fixed start-goal distance, and set the block width randomly for 100 independent runs. In the narrow passage problems, we specify 5 different gap heights, and set random positions of the passage for 100 independent runs. We generate 500 random worlds for each 2D and 3D cases for evaluation. Note we use clearance of 3 pixels for 2D random world, zero clearance for 2D center block and 2D narrow passage, and 2 voxels for 3D random world. The default size of 2D planning problems is $224 \times 224$, and the default size of 3D planning problems is $50 \times 50 \times 50$.

\textbf{Metrics.} For the center block problems, we measure the number of iterations to reach within a path cost threshold, which is some percentage above the optimal cost. For the narrow passage problems, we measure the number of iterations to find a path through the passage. For the random world problems, we examine the iterations each planner spends on finding the initial solution, and path cost improvement after certain numbers of iterations.

\begin{figure*}[hbt!]
    \centering
    \includegraphics[width=\linewidth]{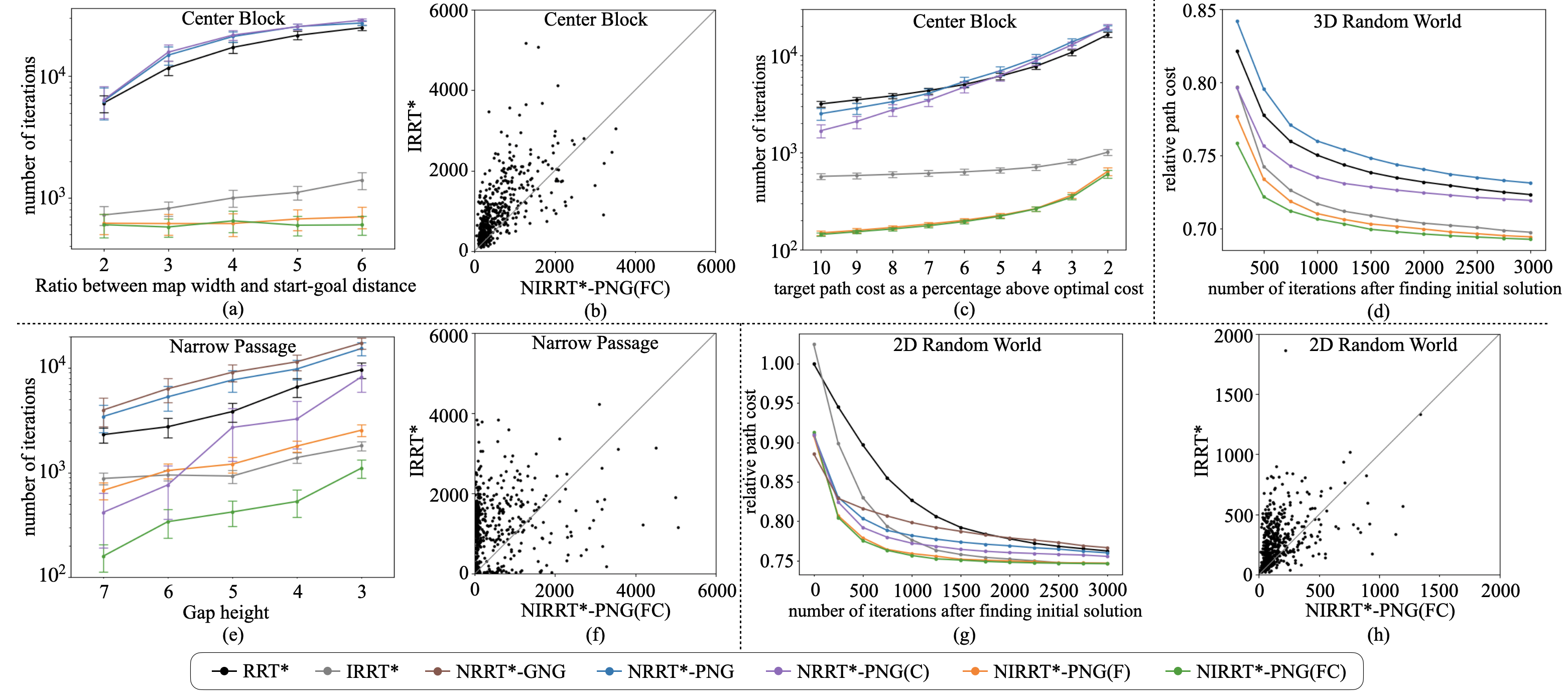}
    \caption{Experiment results. \textbf{Center block}: (a) The average number of iterations to find a path within 2\% of the optimal cost for different map widths; (b) Comparison of the number of iterations IRRT* and NIRRT*\mbox{-}PNG(FC) take to find a path within 2\% of the optimal cost for each center block problem; (c) The average number of iterations to find a path within the specified tolerance of the optimal cost. \textbf{3D Random world}: (d) The average path cost relative to the initial solution of RRT* at different numbers of iterations after finding an initial solution. \textbf{Narrow passage}: (e) The average number of iterations to find a path better than flanking the obstacle for different gap heights; (f) Comparison of the number of iterations IRRT* and NIRRT*\mbox{-}PNG(FC) take to find a path better than flanking the obstacle for each narrow passage problem. \textbf{2D Random world}: (g) The average path cost relative to the initial solution of RRT* at different numbers of iterations after finding an initial solution; (h) Comparison of the number of iterations IRRT* and NIRRT*\mbox{-}PNG(FC) to find an initial solution for each random world problem. Error bars denote 95\% confidence interval. The error bars are not plotted for random worlds for clarity of figures. NRRT*\mbox{-}GNG is not implemented for center block and random world 3D due to incompatible grid sizes and incompatible number of dimensionality.}
    \label{fig:experiment_results}
    \vspace{-10pt}
\end{figure*}


\begin{figure}[hbt!]
    \centering
    \includegraphics[width=\linewidth]{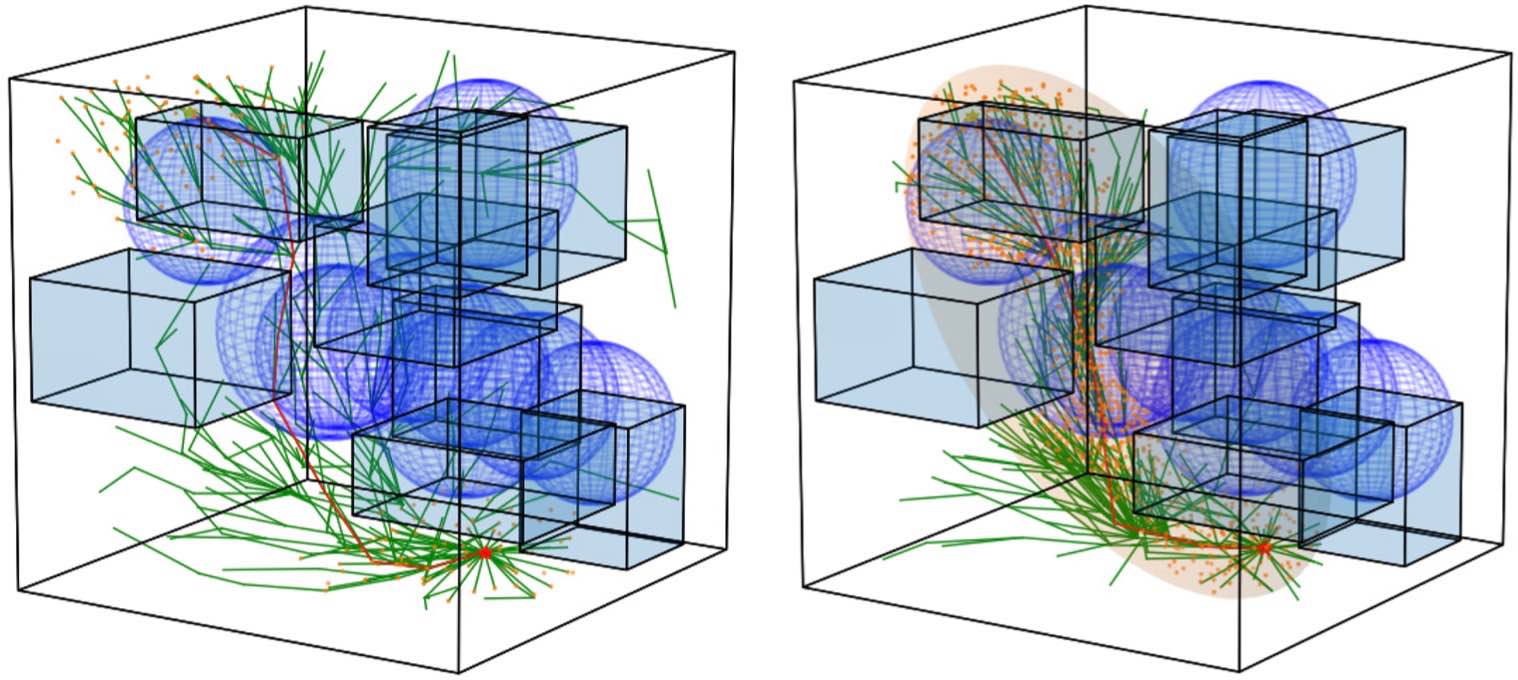}
    \caption{Visualization on planning in 3D random world at 500 iterations. Left: NRRT*\mbox{-}PNG. Right: NIRRT*\mbox{-}PNG(FC).}
    \label{fig:plan-3d}
    \vspace{-20pt}
\end{figure}

\textbf{Baselines.} We compare NIRRT* to RRT*~\cite{karaman2011sampling}, IRRT*~\cite{gammell2014informed}, NRRT*\mbox{-}GNG~\cite{wang2020neural}, and variants of our method across the experiments. NIRRT*\mbox{-}PNG(FC) is our complete algorithm, where F is Neural Focus and C is Neural Connect. NIRRT*\mbox{-}PNG(F) removes Neural Connect from the complete version. NRRT*\mbox{-}PNG is Neural RRT* with the point-based network. NRRT*\mbox{-}PNG(C) uses Neural Connect in addition to NRRT*\mbox{-}PNG. We train a U-Net~\cite{ronneberger2015u} with pretrained ResNet50 weights~\cite{he2016deep} for NRRT*\mbox{-}GNG for 2D problems.

\textbf{Experiment Results.} The center block experiments show in Figure~\ref{fig:experiment_results}(b) that NIRRT*\mbox{-}PNG(FC) outperforms IRRT* in terms of the speed to find near-optimal paths across different problem sizes. The point-based network is able to infer guidance states from the informed subset which are the most promising to converge the path solution to optimum. Both Figure~\ref{fig:experiment_results}(a)(c) show that NIRRT*\mbox{-}PNG(FC) and NIRRT*\mbox{-}PNG(F) have similar performance. The informed subset effectively constrains the region of the point cloud to be around the center block, and significantly simplifies the task of guidance state inference. Therefore, the point-based network performs well even without Neural Connect. Similar to the claim by \cite{wang2020neural} that initial path solution cost of NRRT* is better than RRT*, we see in Figure~\ref{fig:experiment_results}(c) that NRRT*\mbox{-}PNG and NRRT*\mbox{-}PNG(C) are faster than RRT* in terms of reaching within a more relaxed threshold above optimal cost such as 7-10\%. However, the convergence speeds of NRRT* variants tend to be slow and are often worse than RRT* when approaching a tighter bound such as 2-4\%. In contrast, NIRRT* variants work consistently better than IRRT* across thresholds of optimal cost, since the informed subset allows the point-based network to provide finer distribution of the guidance states to continuously refine the path towards the optimal solution.

In the narrow passage setting, NIRRT*\mbox{-}PNG(FC) finds a difficult path through the passage faster and more frequently than IRRT*, as represented in Figure~\ref{fig:experiment_results}(f). The convergence speed of NIRRT*\mbox{-}PNG(FC) outperforms all baselines as shown in Figure~\ref{fig:experiment_results}(e). NIRRT*\mbox{-}PNG(F) performance is similar to IRRT* because the guidance state set usually ends up separated on left and right sides of the gap without Neural Connect, whereas NIRRT*\mbox{-}PNG(FC) is able to connect the guidance state set together through the gap, which helps sampling critical states inside the gap. Both NRRT*\mbox{-}GNG and NRRT*\mbox{-}PNG are worse than RRT*, but NRRT*\mbox{-}PNG(C) works consistently better than RRT*, which indicates the effectiveness of Neural Connect in planning problems with critical states. Note we collect the training dataset for point-based network with optimal paths which requires clearance of 3 pixels, which is equivalent to 7 pixels of the gap height. Figure~\ref{fig:experiment_results}(e) demonstrates that our point-based network generalizes well to planning problems with clearances tighter than training distribution by Neural Focus and Neural Connect, while the CNN model is sensitive to clearance~\cite{wang2020neural}.

For each random world problem, we record the cost of path solution at certain number of iterations after the initial solution is found, and plot these costs relative to the cost of the initial path solution from RRT*. The 2D and 3D results are presented in Figure~\ref{fig:experiment_results}(g) and (d) respectively, and planning in 3D random worlds are visualized in Figure~\ref{fig:plan-3d}. We observe NRRT*\mbox{-}GNG has the best initial solution in 2D since the grid representations are denser than point representations in terms of the whole state space. However, NIRRT* variants converge faster due to continuous improvement of the guidance states. We find that both Neural Connect and Neural Focus contribute to improvement of convergence speed in both 2D and 3D cases. Figure~\ref{fig:experiment_results}(h) shows that NIRRT*\mbox{-}PNG(FC) is faster than IRRT* in terms of finding initial path solution. 

\subsection{Real World Deployment}
We deploy our method and the model trained in 2D random world to a TurtleBot 2i. The demonstration of real world navigation with static obstacles and dynamic humans is available at \url{https://sites.google.com/view/nirrt-star}.

%% file: 05_Conclusions.tex
\section{Conclusions}\label{sec:conclusion}
We present Neural Informed RRT* approach to accelerate optimal path planning by incorporating a point-based network into Informed RRT* for guidance state inference. We introduce Neural Focus to naturally bridge the point-based network and the informed sampling strategy with point cloud representations of free states. We propose Neural Connect to improve quality of the inferred guidance state set by enforcing connectivity. Our simulation experiments show that Neural Informed RRT* outperforms RRT*, Informed RRT*, and Neural RRT* in terms of convergence rate towards optimal solutions in planning problems with varying sizes, critical states, and randomized complicated patterns.

In future work, we want to study how to further improve our algorithm when the planning problem sizes are significantly different from the training distribution. We would like to explore the effectiveness of our work in higher-dimensional problems. It is also interesting to study if we can denoise the guidance state set inferred by the point-based network to offer an end-to-end option for generating feasible and near-optimal paths~\cite{ho2020denoising, janner2022planning}.